\documentclass{article}
\usepackage{spconf,amsmath,epsfig}
\usepackage{tikz}
\usepackage{multirow}
\usepackage{tabularx}
\usepackage{pifont}
\usepackage{hyperref}
\usepackage{bm}

\newcommand{\m}[1]{{\mbox{{\fontencoding{T1}\sffamily\slshape{#1\/}}}}}
\renewcommand{\d}[1]{{\mbox{\boldmath$#1$}}}

\newcommand*\circled[1]{\tikz[baseline=(char.base)]{\node[shape=circle,draw,inner sep=1pt] (char) {#1};}}

\title{Deep Self-taught Learning for Remote Sensing Image Classification}
\name{Anika Bettge, Ribana Roscher, Susanne Wenzel}
\address{Remote Sensing, Institute of Geodesy and Geoinformation, University of Bonn, Germany, 53115 Bonn\\
s7anbett@uni-bonn.de, 
ribana.roscher@uni-bonn.de, 
wenzel@igg.uni-bonn.de}
\begin{document}
%
\maketitle
\begin{abstract}
This paper addresses the land cover classification task for remote sensing images by deep self-taught learning. 
Our self-taught learning approach learns suitable feature representations of the input data using sparse representation and undercomplete dictionary learning.
We propose a deep learning framework which extracts representations in multiple layers and use the output of the deepest layer as input to a classification algorithm. 
We evaluate our approach using a multispectral Landsat 5 TM image of a study area in the North of Novo Progresso (South America) and the Zurich Summer Data Set provided by the University of Zurich.
Experiments indicate that features learned by a deep self-taught learning framework can be used for classification and improve the results compared to classification results using the original feature representation. 
\end{abstract}
\begin{keywords}
self-taught learning, deep learning, archetypal analysis, landcover classification, remote sensing
\end{keywords}
\section{Introduction}
\label{sec:intro}
Classification of remote sensing images is an important task for land cover mapping. 
Recently, deep learning has become a valuable approach particularly for such classification tasks. 
As already pointed out by \cite{bengio2013representation}, the most successful approaches are supervised deep learning frameworks using a huge amount of labeled data for training. 
However, labeled data are scarce for remote sensing applications. 
In contrast, huge amounts of unlabeled data are available and easy to acquire.

In our approach we use self-taught learning (STL, \cite{raina2007self}) which has turned out as a valuable procedure to the combined exploitation of unlabeled and labeled data, without the constraint that both datasets need to follow the same distribution. 
Therefore, we can utilize datasets from further scenes and acquisition times for feature/representation learning. 

The most common approach to STL is sparse representation (SR), which learns features in an unsupervised way in order to use them for supervised classification.
Some approaches use deep sparse representations DSR, which shows improved results over shallow representations.
E.g., \cite{dsn} propose a deep unsupervised feature learning approach, but include only labeled data. 
The authors of \cite{kemker2017self} use stacked convolutional autoencoders as well as  independent component analysis with non-linearity for learning deep representations. 
He et al. \cite{he2014unsupervised} also use multi-layers of SR to obtain higher-level features. For this they combine a fully unsupervised feature learning procedure with hand-crafted feature extraction and pooling. 
The deep belief network of \cite{lin2010deep} benefits from the geometric data structure achieved by the local coordinate coding. They represent all data samples in two stacked layers as sparse linear combination of anchor points.
However, so far all deep feature learning approaches do not produce fully interpretable representations. 

In this paper, the overall goal is to learn deep features with the help of big amounts of unlabeled data, leading to good classification results and interpretable features, the latter being important for many classification or unmixing tasks \cite{romer2012early}. 
We achieve this by designing a deep framework, called deep STL (DSTL), which combines STL and deep learning concepts.  
We extend  the shallow approach of \cite{roscher2015landcover}, which shows that high classification accuracies can be achieved by combining STL with archetypal dictionaries.
Furthermore, we use the approach of \cite{thurau2010yes} to find archetypes (extreme points of the data distribution), and adapt the dictionary learning to be suitable for our deep learning framework.



\section{Deep Self-Taught Learning}
\label{sec:DSN}
STL uses unlabeled data $^u\!\m X = [^u\!\d x_q], q = 1, ..., Q$, training data  $^{tr}\!\m X = [^{tr}\!\d x_n], n = 1, ..., N$ with labels $^{tr}\!\d y = \left[ ^{tr}\!y_n\right] $ given, and test data $^t\!\m X = [^{t}\!\d x_p], p = 1, ..., P$., also with labels $^{t}\!\d y$. All data samples consist of $M$-dimensional feature vectors $\d x\in {\rm I\!R}^M$, and the labels $y \in \lbrace 1, ..., c, ..., C\rbrace$, where $C$ is the number of classes. The labels are also represented by target vectors $\d t = \left[ t_c\right] $ of length $C$ coding the label with $t_c = 1$ for $y=c$ and $t_c = 0$ otherwise.

We use SR approximating each sample by a linear combination of only a few elements of a dictionary. 
We initialitize the dictionary following archetypal analysis \cite{cutler1994archetypal}, achieving a sparse data approximation $\d x_n \approx \m D~ \boldsymbol \alpha_n$, with an ($M \times K$)-dimensional dictionary $\m D$. 
To learn the dictionary $\m D = [\d d_k]$ we apply simplex volume maximization (SiVM) \cite{thurau2010yes}. 
This approach finds archetypes as extreme points lying on the convex hull of the unlabeled data set $\lbrace\d d_k\rbrace \in \lbrace ^u\!\d x_q\rbrace$, where $K\leq Q$. 
The coefficient vectors $\boldsymbol \alpha_n$ are the new SR of the data samples. 
We derive these SRs by minimizing $\Vert \m D \boldsymbol \alpha_n - \d x_n \Vert$ subject to non-negativity constraint $ \alpha_{kn} \geq 0 $ for all $k$ and sum-to-one constraint $\sum_{k=1}^K \alpha_{kn} = 1$.  

\begin{figure}[t]
\begin{minipage}[t]{1.0\linewidth}
  \centering
 \centerline{\epsfig{figure=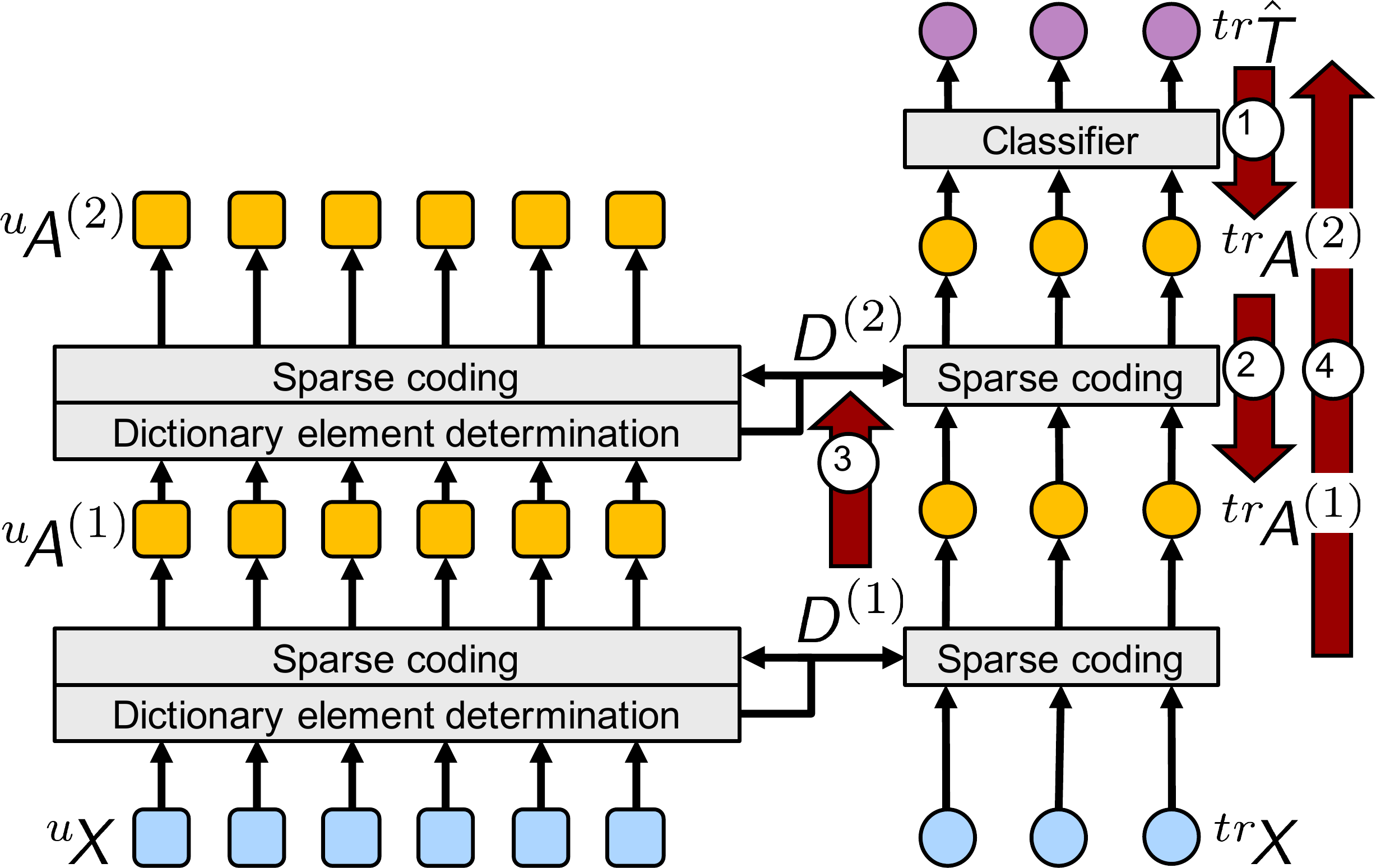,width=8.5cm}}
\end{minipage}
\caption{Structure and update procedure of the DSTL approach:
(left block) unlabeled and (right block) labeled data with classifier. The numbers represent the update procedure: \ding{172} gradient descent update of the $L^{th}$ training representation; \ding{173} backpropagation of the gradient; \ding{174} dictionary update with training representations; \ding{175} learning new training representations and classifier.}
\label{fig:dsn}
\end{figure}

We extend the STL approach to DSTL to learn deep representations. 
Our network structure is shown in Fig.~\ref{fig:dsn}. 
The left side illustrates the exploitation of unlabeled data, and the right side the used of the labeled data; the data and their representations are symbolized by filled rectangle for the unlabeled and by circles for the labeled data. 
In general the network consists of $L$ layer, where Fig. \ref{fig:dsn} shows the structure for $L=2$.
Pre-training is performed layer-wise, so that in each layer $l = 1, ..., L$ the dictionary elements $\m D^{(l)}$ are determined from $^u\! \m X$ for $l = 1$ and from $^u\!\m A^{(l-1)} = \left[ ^u\!\boldsymbol \alpha_q^{(l-1)} \right] $ for $l \neq 1$. 
Given the dictionaries, we learn $^{u}\!\m A^{(l)}$ by least square estimation with non-negativity and sum-to-one constraint (left side of Fig. \ref{fig:dsn}). 
Likewise, we learn $^{tr}\!\m A^{(l)}$ of the labeled data of the network from $\m D^{(l)}$. 
We train the logistic regression model (\cite{bishop2006pattern} p. 205-210) with the new features $^{tr}\!\m A^{(L)}$ predicting conditional probabilities $P(c|^{tr}\!\d x_n)$, which we can interpret as $^{tr}\!\hat{\m T} = [^{tr}\!\hat{\m t_n}]$.

To improve the network the reconstruction error can be minimized by updating the networks parameter.
The number of parameters which need to be learned in a 2-layers network are  \mbox{$(M + K^{(2)})\cdot K^{(1)} + (K^{(2)}+1)\cdot C$} with $K^{(\cdot)}$ being the number of dictionary elements in the layer $(\cdot)$. It contains the number of dictionary entries in each layer and parameters of the classifier models.
In order to learn these parameters we perform the following steps illustrated by \circled{1} - \circled{4} in Fig. \ref{fig:dsn}: 
In the first update step the dictionaries are fixed and only the training representations are updated.
Given $^{tr}\!\d t_n$ and $^{tr}\! \hat{\d t}_n$, 
the backpropagation loss function is given by the following equation:
\begin{equation}
J\left( ^{tr}\!\boldsymbol \alpha_n^{(L)}\right)  = \frac{1}{2} \Vert ^{tr}\!\d t_n - ^{tr}\! \hat{\d t}_n \Vert^2 .
\end{equation}
Here the target vectors $^{tr}\!\d t_n$ expresses the true membership of the training samples to the classes in the from of the 1-of-C coding scheme, and the  $^{tr}\! \hat{\d t}_n$ is the likewise encoded conditional probability for class membership estimated by our network.
With the help of the gradient of this loss function 
with respect to the training representations $^{tr}\!\boldsymbol \alpha_n^{(L)}$ we update the training representations of the last layer:
\begin{equation}
^{tr}\!\alpha_{kn}^{*(L)} = ^{tr}\!\alpha_{kn}^{(L)} - a \frac{\partial J(^{tr}\!\boldsymbol \alpha_n^{(L)})}{\partial  ~^{tr}\!\alpha^{(L)}_{kn}}.
\end{equation} 
 Here the gradient is clipped element-wise to a threshold $t_1$ to avoid too excessive modifications of the representations \cite{Goodfellow-et-al-2016}.
We then backpropagate the gradient through the net in order to update all labeled representations using
\begin{equation}
^{tr}\!\m A^{*(l)} = \m D^{(l+1)} ~^{tr}\!\m A^{*(l+1)} ,
\end{equation}
for $l = L-1, ..., 1$ (\circled{1} and \circled{2}).

In step \circled{3}, given the updated training representations, we update the dictionaries $\m D^{(l)}$ using the gradient descent method as proposed by \cite{dsn}. 
To compute the dictionary update, we define a loss function
\begin{equation}
J_D\left( \m D^{(l)}\right)  = \frac{1}{2} \Vert \m D^{(l)}~ ^{tr}\!\m A^{*(l)} - ^{tr}\!\m A^{*(l-1)}\Vert^2,
\end{equation}
which will be minimized.
In the first layer, $^{tr}\!\m A^{*(0)}$ is given by the original data $^{tr}\!\m X$.
The gradient descent updating rule for the $k^{th}$ dictionary element of the $l^{th}$ layer is given by 
\begin{equation}
\d  d_k^{*(l)} = \d d_k^{(l)}- \gamma \left( \m D^{(l)}~ ^{tr}\!\m A^{(l)} - ^{tr}\!\m A^{*(l-1)} \right)~ ^{tr}\!\boldsymbol \alpha_k^{(l)},
\end{equation}
where $\gamma$ is the learning rate. Again the gradient 
is clipped to a threshold $t_2$.
Due to the dictionary updates their entries do not represent raw data samples anymore, which makes them not interpretable. 
We want to keep the dictionary elements interpretable by restricting them to true data samples. 
In case a dictionary element has changed sufficiently, we shift it to the nearest neighbor in feature space which contains the set of unlabeled data samples.
Step \circled{4} finally readjusts the labeled representations with the updated $\m D^{(l)}$ by minimizing the reconstruction error of $^{tr}\!\m X$ and updates the classifier. 
We iterate steps \circled{1} - \circled{4} until convergence of the dictionaries.

\section{Experimental Setup And Results}
\label{sec:ExperimentalSetupAndFirstResults}

\begin{figure*}[t]
\begin{minipage}[t]{1.0\linewidth}
  \centering
 \centerline{\epsfig{figure=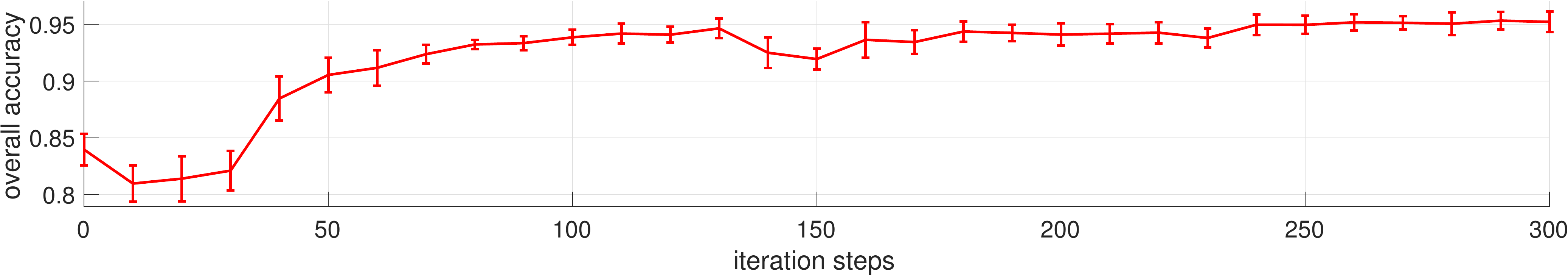,width=\textwidth}}
\end{minipage}
\caption{Average and standard deviation over 10 runs of overall accuracy [\%] of the DSTL approach for the Landsat 5 Data Set (right) over up to $300$ iterations..}
\label{fig:results_cZurich}
\end{figure*}
In this section we test our DSTL approach for two multi-spectral image data sets. 
For this we apply our two layered DSTL to the data sets and compare the accuracy with the results of a simple logistic regression.
In Section \ref{sec:dataSets} the two data sets are briefly introduced, followed by the data (Sec. \ref{sec:dataSetup}) and experimental setup (Sec. \ref{sec:expSetup}). Finally, the results are presented in Sec. \ref{sec:results}.
\subsection{Data Sets}
\label{sec:dataSets}
We use the following two multi-spectral data sets for the testing of our approach:
\vspace{0.1cm}\\ \textbf{Landsat 5 Data Set: } Our first data set is a multi-spectral Landsat 5 TM image from a study area located in the North of Novo Progresso (South America). 
It contains data for \mbox{$6$ bands} (red, green, blue, Mid-Infrared and two NIR-bands) for $6,962 \times 7,921$ pixels. 
Parts of the image are labeled (approx. 57,000 pixels) with 6 classes (see Tab. \ref{tab:landsat5}). 
  Additionally, we collect about $600,000$ image patches from diverse areas worldwide as unlabeled data samples.
\vspace{0.1cm}\\ \textbf{Zurich Summer Data Set: }
The Zurich Summer Data Set, provided by the University of Zurich 
\cite{volpi2015semantic}, contains 20 VHR images of Zurich recorded 2002 by the QuickBird satellite.
The images comprise four bands: red, green, blue and NIR with spatial resolution of $61.5~cm$. They are labeled by 8 urban and periurban classes (roads, buildings, trees, grass, bare soil, water, railways, and swimming pools).

\subsection{Data Setup}
\label{sec:dataSetup}
For both data sets we choose $5 \times 5$-pixel image patches, leading to $100$-dimensional input feature vectors for the Zurich Summer Data Set and $150$-dimensional input feature vectors for the Landsat 5 Data Set.
These input vectors are global contrast normalized and then shifted to positive values as input vectors for the DSTL network.
\vspace{0.1cm}\\ \textbf{Landsat 5 Data Set: }
We randomly extract ten sub-data sets with $1,000$ training samples ($^{tr}\!\m X$), around $56,000$ test samples ($^{t}\!\m X$), and $1,000$ validation samples each from the Landsat 5 image. We use the patches (around $600,000$ pixels) as unlabeled samples ($^{u}\!\m X$).
\vspace{0.1cm}\\ 
\textbf{Zurich Summer Data Set: }
We randomly extract $500$ validation samples from one of the 20 images, the rest of the labeled data of this image is used as test samples ($^{t}\!\m X$), and $1,000$ training samples ($^{tr}\!\m X$) are selected from the remaining 19 images.
The $10,000$ unlabeled ($^{u}\!\m X$) samples are randomly selected from all images. The data selection is done 20 times, so that from each image test and validation data are selected.

\subsection{Experimental Setup}
\label{sec:expSetup}
In our experiments we create a DSTL with two layers to test if the test accuracy benefits from the DSTL approach over a simple logistic regression.
We perform the following experiments on the two data sets:
\vspace{0.1cm}\\ \noindent\textbf{Landsat 5 Data Set: }  The DSTL approach is carried through with $20$ archetypes in the first layer and $30$ in the second. 
\vspace{0.1cm}\\ \textbf{Zurich Summer Data Set: } The DSTL approach is performed with $30$ and $40$ archetypes for the two layers.
\vspace{0.1cm}

In all experiments the threshold of the gradient clipping is set to  $t_1 = t_2 = 0.001$ and the learning rates to $ a = \gamma = 1$. 
The DSTL update 
is iterated $1,000$ times to find the best dictionary, judged by application to the validation data.
We achieve the best results, in terms of classification accuracy, by stacking the representations of all layers for classification, similar to the idea used in denseNet \cite{huang2016densely}.
\subsection{Results}
\label{sec:results}
In all our experiments we achieve an improvement over the original representations: \vspace{0.1cm}\\
\textbf{Landsat 5 Data Set: }
Table \ref{tab:landsat5} shows the class-wise, overall, and average test accuracy as well as the Kappa coefficient for the experiment of the Landsat 5 Data Set. 
Our DSTL approach with stacked representations yields better overall and average accuracies than the original data, but the Kappa coefficient is decreased. The average and standard deviation of the overall accuracy is illustrated in Fig. \ref{fig:results_cZurich}. It becomes obvious, that the average increases over the most iterations and the standard deviation is with maximal $0.03$ \% very small.
\begin{table}[t!]
\caption{Class-wise accuracies [\%], overall accuracy [\%], average accuracy [\%] and Kappa coefficient (Kappa) obtained by logistic regression using the original features of the Landsat 5 Data Set, and logistic regression on the stacked deep representations of the DSTL approach. The average results over ten runs is given with the standard deviation. The best results are highlighted in bold-print. }
\begin{minipage}[t]{0.5\textwidth}
  \centering
 \begin{tabular}{l c c }
 \hline
 &  \multicolumn{1}{c}{original features} & \multicolumn{1}{c}{DSTL features}	\\  \hline
 water &$ 90.0 \pm 23.1$ & \bm{$97.9 \pm 2.0$} \\ 
urban & $88.9 \pm 5.1$ &\bm{$91.5 \pm 4.8$} \\ 
secondary forest&$55.4 \pm 7.9$ &\bm{$71.1 \pm 3.9$} \\ 
pasture &\bm{$99.1 \pm 0.8$} &$98.5 \pm 1.3$ \\ 
burned pasture &\bm{$100.0 \pm 0.0$} &$100.00 \pm 0.1$ \\ 
primary forest & \bm{$99.9 \pm 0.1$} &$99.5 \pm 0.2$ \\ 
 \hline overall & $94.2 \pm0.9$ &\bm{$95.9 \pm 0.4$} \\ 
average & $ 88.9 \pm 4.2$ &\bm{$93.3 \pm 1.1$} \\ 
Kappa &\bm{$0.86 \pm 0.02$} &$0.90 \pm 0.01$ \\ \hline 
\end{tabular} 
\end{minipage}
\label{tab:landsat5}
\end{table}
\vspace{0.1cm}\\ \textbf{Zurich Summer Data Set: } 
Table \ref{tab:zurich} shows that the DSTL with stacked features leads to an improvement in the mean over-all and mean average accuracy. Here also the kappa coefficient increases. 
\begin{table}
\centering
\caption{Mean results of the logistic regression of the original Zurich Summer Data Set and of the stacked deep representations of the DSTL approach.}
\begin{tabular}{l c c}
\hline
 & \multicolumn{1}{c}{original features} & \multicolumn{1}{c}{DSTL features} \\ \hline
overall accuracy& $61.6 \pm 0.7$ \% & $64.7 \pm 1.3$ \%\\
average accuracy & $49.7 \pm 1.5$ \% & $55.0 \pm 1.6$ \%\\
Kappa& $0.46 \pm 0.01$& $0.50 \pm 0.02$ \\ \hline
\end{tabular}
\label{tab:zurich}
\end{table}
Running this experiment with $1,000$ iterations with Matlab version 16b on an Intel Core i5-2400 processor takes ca. 14 hours.
\vspace{0.1cm}

 We expect to further improve the results by using larger dictionaries, but this will significantly increase run time. 
\section{Conclusion}
\label{sec:conslusion}
In this work we present a deep self-taught learning framework to combine the advantages of the STL with interpretable dictionaries and the deepness of neural networks. The accuracy is tested with two different multi-spectral image data sets. Further research will deal with a deeper DSTL network and interpretable non-linearity to raise the the accuracy. 
Altogether, the deep self-taught learning framework profits by the huge amount of unlabeled data, so that the learned deep features improve the results compared to classification results using the original feature representation and are achieved with still interpretable dictionaries. 

\section*{Acknowledgments}
This work has partly been supported  by the EC under contract number H2020-ICT-644227-FLOURISH.

\vfill

\bibliographystyle{IEEEbib}
\bibliography{references}

\end{document}